\documentclass[10pt,journal,compsoc]{IEEEtran}

\ifCLASSOPTIONcompsoc
  \usepackage[nocompress]{cite}
\else
  \usepackage{cite}
\fi
\usepackage{microtype}
\usepackage{graphicx}
\usepackage{comment}

\ifCLASSOPTIONcompsoc
  \usepackage[caption=false,font=footnotesize,labelfont=sf,textfont=sf]{subfig}
\else
  \usepackage[caption=false,font=footnotesize]{subfig}
\fi

\usepackage{booktabs} %
\usepackage{combinedgraphics}
\usepackage{multirow}

\usepackage[utf8]{inputenc}
\usepackage{amsmath}
\usepackage{amssymb}

\usepackage{mathtools}
\usepackage{color}
\usepackage{amsthm}
\usepackage{dsfont}
\usepackage{bbm}

\usepackage{algorithm}
\usepackage{algpseudocode}

\usepackage{dcolumn}
\newcolumntype{d}[1]{D{.}{.}{#1}}

\usepackage{colortbl}

\usepackage{hyperref}

\usepackage{url}

\usepackage{xcolor}
\definecolor{DSgray}{cmyk}{0,1,0,0}

\newtheorem{definition}{Definition}

\DeclareMathOperator*{\argmin}{argmin}

\usepackage{enumitem}
\setenumerate[1]{itemsep=2pt,partopsep=2pt,parsep=0pt,topsep=2pt}
\setitemize[1]{itemsep=2pt,partopsep=2pt,parsep=0pt,topsep=2pt}

\begin{document}

\title{Isolation Kernel: The X Factor in Efficient and Effective Large Scale Online Kernel Learning}

\author{Kai~Ming~Ting,
        Jonathan~R.~Wells~,
        and~Takashi~ Washio%
\IEEEcompsocitemizethanks{\IEEEcompsocthanksitem K. M. Ting and J. R. Wells are with Federation University\protect\\
Corr. e-mail: \url{kaiming.ting@federation.edu.au}
\IEEEcompsocthanksitem T. Washio is with Osaka University}%
}

\IEEEtitleabstractindextext{%
\begin{abstract}
Large scale online kernel learning aims to build an efficient and scalable kernel-based predictive model incrementally from a sequence of potentially infinite data points. A current key approach focuses on ways to produce an {\em approximate finite-dimensional feature map}, assuming that the kernel used has a feature map with intractable dimensionality---an assumption traditionally held in kernel-based methods. While this approach can deal with large scale datasets efficiently, this outcome is achieved by compromising predictive accuracy because of the approximation.
We offer an alternative approach which overrides the assumption and puts the kernel used at the heart of the approach. It focuses on creating an {\em exact, sparse and finite-dimensional feature map} of a kernel called Isolation Kernel. Using this new approach, to achieve the above aim of large scale online kernel learning becomes extremely simple---simply use Isolation Kernel instead of a kernel having a feature map with intractable dimensionality. We show that, using Isolation Kernel, large scale online kernel learning can be achieved efficiently without sacrificing accuracy.

\end{abstract}

\begin{IEEEkeywords}
Data dependent kernel, online kernel learning, kernel functional approximation, large scale data mining
\end{IEEEkeywords}}

\maketitle

\IEEEraisesectionheading{\section{Introduction}\label{sec:introduction}}

\IEEEPARstart{I}{n} the age of big data, the ability to deal with large datasets or online data with potentially infinite data points is a key requirement of machine learning methods. Kernel methods are an elegant machine learning method to learn a nonlinear boundary from data. However, its applications in the age of big data is limited because of its perennial problem of high computational cost on high dimensional and large datasets. 

Current state-of-the-art large scale online kernel learning focuses on improving efficiency. There are two key current approaches (e.g., \cite{BudgetOnline-2010,BudgetSGD-2012,BudgetPerceptron2007, LargeScaleOnlineKernelLearning}) to gain efficiency  through approximation, i.e., (i) limiting the number of support vectors, and (ii) using an approximate feature map. These approaches assume that a kernel used has a feature map with intractable dimensionality---an assumption traditionally held in kernel-based methods. While successfully achieving the efficiency gain, both approaches must also manage the inevitably negative impact of the approximation  on accuracy as much as possible. 

Here we offer a third approach which  overrides the assumption and puts the kernel used at the heart of the approach. The idea is to create and use an {\em exact, spare and finite-dimensional} feature map of a kernel called Isolation Kernel \cite{ting2018IsolationKernel,IsolationKernel-AAAI2019}. We show that a feature map with these characteristics yields an online kernel learning method which achieves efficiency gain without the need to manage accuracy degradation. %

For example, the predictive accuracy of a current method can be degraded to an unacceptable low level when it is applied to datasets having more than 1000 dimensions; while the new method can maintain high accuracy with equivalent or better efficiency gain. The details are provided in Section~\ref{sec_experiments}.

The contributions of this paper are:
\begin{enumerate}
    \item Offering a new approach to online kernel learning which overrides the usual assumption of kernel-based methods, i.e., the kernel has a feature map with intractable dimensionality.
    \item Revealing that a recent Isolation Kernel has an exact, sparse and finite-dimensional feature map.
    
    \item Showing that Isolation Kernel with its exact, sparse and finite-dimensional feature map is a crucial factor in  enabling efficient large scale online kernel learning 
  without compromising accuracy.  
    Specifically, the proposed feature map enables three key elements: learning with exact feature map, efficient dot product and GPU acceleration that lead to the success of the proposed method.
    \item Demonstrating the impact of Isolation Kernel on an existing algorithm of online kernel learning called Online Gradient Descent (OGD) and also support vector machines (SVM). Using Isolation Kernel, instead of a kernel with a feature map with intractable dimensionality, the same algorithms (OGD and SVM) often achieve better predictive accuracy and always have significantly faster runtime by up to three orders of magnitude.\\
    Compare with existing online kernel learning methods, (a) OGD with Isolation Kernel has better accuracy than the state-of-the-art online OGD called NOGD (Nystr\"{o}m Online Gradient Descend \cite{LargeScaleOnlineKernelLearning}); and it runs up to one order of magnitude faster. (b) SVM with Isolation Kernel have similar or better accuracy than SVM with $\chi^2$ additive kernel \cite{AdditiveKernel-PAMI2012}; and it runs up to one order of magnitude faster in high dimensional datasets.
    \item Unveiling for the first time that (a) the Vonoroi-based implementation of Isolation Kernel produces better predictive accuracy than the tree-based implementation in kernel methods using OGD; (b) the GPU version of the implementation is up to four orders of magnitude faster than the CPU version.
\end{enumerate}

 The rest of the paper is organised as follows. Section~\ref{sec_challenges+approach} describes the current challenge and key approach in large scale online kernel learning. Section \ref{sec_advantages} presents the two previously unknown advantages of Isolation Kernel and its current known advantage. Section \ref{sec_IK} describes the current understanding of Isolation Kernel: its definition, implementations and characteristics. Section \ref{sec_learning_IK} presents our four conceptual contributions in relation to learning with exact feature map of Isolation Kernel. Its applications to online gradient descent and support vector machines are presented in Section \ref{sec_applications}. The experimental settings and results are provided in the next two sections. Section \ref{sec_relation} describes the relationship with existing approaches to efficient kernel methods, followed by discussion and concluding remarks in the last three sections.

\section{Current challenges and key approach in Large Scale online kernel learning}
\label{sec_challenges+approach}
We will describe the current challenges in online kernel learning and an influential approach to meet one of the challenges in the next two subsections

\subsection{Challenges in online kernel learning}
\label{sec_challenges}

Kernel methods are an elegant way to learn a nonlinear boundary. 
But they are hampered by high computational cost. There are two approaches in improving its efficiency, depending on whether one is solving the dual or primal optimisation problem.

First, in solving the dual optimisation problem that employs the kernel trick to avoid feature mapping, one of its main computational costs is due to the prediction function used, i.e.,
$_{dual}f({\bf x}) = \sum_{i=1}^s \alpha_i c_i K({\bf 
 x}_i,{\bf x})$, where $K$ is the chosen kernel function; $\alpha_i$ is the learned weight and $c_i \in \{+,-\}$ is the class label of support vector ${\bf x}_i$; and $s$ is the number of support vectors. The sign of $_{dual}f({\bf x})$, i.e., $+$ or $-$, yields the final class prediction.

The evaluation of the prediction function $_{dual}f$ has high cost if the number of support vectors is high.

The first approach to improve efficiency is to limit the number of support vectors, and it is often called budget online kernel learning (e.g., \cite{BudgetPerceptron2007,BudgetOnline-2010,BudgetSGD-2012}; and see \cite{LargeScaleOnlineKernelLearning} for a review.) The key limitation of this approach is that it is unable to deal with an  unlimited number of support vectors.

Second, abandoning the kernel trick by using an approximate feature map of a chosen nonlinear kernel,
one usually solves the primal optimisation problem because its prediction function has less cost.
The evaluation of the prediction function $_{primal}f$ has cost proportional to the number of features in the feature map $\Phi$, i.e., $_{primal}f({\bf x}) = \left<{\bf w}, \Phi({\bf x})\right>$, where ${\bf w} = \sum_{i=1}^s \alpha_i c_i \Phi({\bf x}_i)$ can be pre-computed once the support vectors are determined. 

The success of this second approach relies on a method to produce a good approximate feature map. This approximation can be costly; and some method can only afford to use a data subsample for the approximation in order to reduce the time complexity. 
This requirement has the same impact of degrading the accuracy as limiting the number of support vectors in $_{dual}f$ used in the first approach.

Kernel methods, that are aimed for large scale datasets, solve the primal optimisation problem because  $_{primal}f$ has constant time cost, independent of the number of support vectors, as used in a recent system \cite{LargeScaleOnlineKernelLearning}.

In a nutshell, the key challenge in large scale online kernel learning that employs $_{primal}f$ is to
\textbf{obtain a good approximate feature map of a chosen nonlinear kernel function}
such that the inevitable negative impact they have on accuracy is reduced as much as possible.

\subsection{An existing influential approach}

The need to approximate a feature map of a chosen nonlinear kernel arises because existing nonlinear kernels such as Gaussian and polynomial kernels have either infinite or a large number of features.
Table \ref{tab:feature_size} provides the sizes of their feature maps. 

\begin{table}[!t]
 \centering
 \caption{Feature map size comparison for three kernels. $C(\cdot,\cdot)$ is a binomial coefficient. See \cite{Polykernel-Chang:2010} for details about polynomial kernel. ${\bf x}$ and ${\bf y}$ are data points of $d$ dimensions; $D$ is the given dataset. All other variables are scalar parameters.}
\label{tab:feature_size}
 \begin{tabular}{lcc}
  \toprule
Kernel & & Feature map (\#dimensions) \\
  \midrule
Gaussian & $exp(-\gamma \parallel {\bf x} - {\bf y} \parallel^2)$ & infinite\\
Polynomial & $(\alpha\ {\bf x} \cdot {\bf y} + r)^h$ & $C(d+h,h)$\\ \midrule
Isolation & space-partitioning($t,\psi | D$) & $t\psi \rightarrow t$\\
  \bottomrule
 \end{tabular}
\end{table}

One influential approach to meet the first key challenge is kernel functional approximation; and its two popular methods are: (a) The Nystr\"{o}m embedding method \cite{Nystrom_NIPS2000} which uses sample points from the given dataset to construct a matrix of low rank $r$ and derive a vector representation of data of $r$ proxy features. (b) Derive random features based on Fourier transform \cite{RandomFeatures2007,OrthonogalRandomFeatures2016} or Laplacian transform \cite{RandomLaplaceFeatures}, independent of the given dataset.
Both produce an approximate feature map of a chosen nonlinear kernel using proxy features which are aimed to be used as input to a linear learning algorithm.

A recent proposal of online kernel learning \cite{LargeScaleOnlineKernelLearning} has employed the  Nystr\"{o}m embedding method to meet the challenge.
The algorithm called NOGD (Nystr\"{o}m Online Gradient Descend) has shown encouraging results, dealing successfully with large scale datasets and has good predictive  accuracy in online setting for datasets less than 800 dimensions \cite{LargeScaleOnlineKernelLearning}.

However, because the feature map is an approximation, the approach reduces the time and space complexities with the expense of accuracy. In addition, we demonstrate that NOGD has performed poorly on datasets more than 1000 dimensions (see results in Section \ref{sec_exp_results}).

We show here that the challenge on online kernel learning only exists because of the kind of kernels employed.
For existing commonly used  nonlinear kernels, the dimensionality of their feature maps is not controllable by a user, and has infinite or a large number of dimensions. 
The kernel functional approximation approach is a workaround without addressing its root cause of the challenge. 

In summary, both two current approaches to large scale online learning have the tacit assumption that the kernel used has a feature map with intractable dimensionality. We show in the next section that this assumption can be overridden to give rise to a new approach.

\section{Advantages of Isolation Kernel}
\label{sec_advantages}
We show here that a recent kernel called
Isolation Kernel \cite{ting2018IsolationKernel,IsolationKernel-AAAI2019} has two advantages, unbeknown previously, compared with existing data independent kernels:
\renewcommand{\labelenumi}{\roman{enumi})}
\begin{enumerate}
    \item The unique characteristic is that Isolation Kernel has an {\em exact} feature map which is {\em sparse} and has a finite number of features that can be controlled by a user. 

\item The sparse and finite-dimensional representation
i.e., each feature vector has exactly $t$ out of the $t\psi$ elements being non-zero, 
enables an efficient dot product implementation.

\end{enumerate}

As Isolation Kernel has no functional form, the feature map is exact in the sense that both the kernel and the feature map are obtained from the same data dependent isolation mechanism and no additional conversion is required.

The first advantage eliminates the need to get an approximate feature map (through kernel functional approximation or other means)---when an exact feature map is available, there is no reason to use an approximate feature map. The existence of exact feature map destroys the premise of the challenge in online kernel learning.

The unique characteristic of Isolation Kernel enables  
kernel learning to solve the primal optimisation problem efficiently. 
This is because evaluating the prediction function can be conducted more efficiently using $_{primal}f({\bf x}) = \left<{\bf w}, \Phi({\bf x})\right>$, where ${\bf w} = \sum_{i=1}^s \alpha_ic_i \Phi({\bf x}_i)$ can be pre-computed once the support vectors are determined. This is applicable in the testing stage as well as in the training stage.

The second advantage  of sparse and finite-dimensional representation enables the dot product in $_{primal}f$ to be computed efficiently, i.e., orders of magnitude faster than that without the efficient implementation under some condition. 

We show that the above advantages of Isolation Kernel allow an efficient kernel-based prediction model to deal with an unlimited number of support vectors in a sequence of infinite data points.

In a nutshell, the type of kernel used, which has infinite or large number of features, has necessitated an intervention step to approximate its feature map. A considerable amount of research effort \cite{Polykernel-Chang:2010,Nystrom_NIPS2000,RandomFeatures2007, RandomLaplaceFeatures} has been invested in order to produce a feature map %
that has a more manageable dimensionality. 
Using the type of kernel such as Isolation Kernel---which has an exact, user-controllable finite-dimensional feature map---eliminates the need of such an intervention step for feature map approximation.

\subsection{One known advantage of Isolation Kernel}
\label{sec_data_dependent_kernels}

In addition to the above two (previously unknown) advantages, Isolation Kernel has one known advantage, i.e., it is data dependent \cite{ting2018IsolationKernel,IsolationKernel-AAAI2019}, as opposed to {\em data independent} kernels such as Gaussian and Laplacian kernels. It is solely dependent on data distribution, requiring neither class information nor explicit learning. Isolation Kernel has been shown to be a better kernel than existing kernels  in SVM classification \cite{ting2018IsolationKernel}, and  has better accuracy than existing methods such as multiple kernel learning \cite{rakotomamonjy2008simplemkl} and distance metric learning \cite{zadeh2016geometric}. Isolation Kernel is also a successful way to kernelise density-based clustering \cite{IsolationKernel-AAAI2019}.

These previous works have focused on the improvements on task-specific performances; but the use of Isolation Kernel has slowed the algorithms' runtimes \cite{ting2018IsolationKernel, IsolationKernel-AAAI2019}. They also have focused on the use of kernel trick, and the feature map of Isolation Kernel was either implicitly stated \cite{ting2018IsolationKernel} or not mentioned at all \cite{IsolationKernel-AAAI2019}.

Here we present the feature map of Isolation Kernel and its characteristic, and the benefits it bring to online kernel learning that would otherwise be impossible---{\em a kernel learning which can deal with infinite number of support vectors}; and run efficiently to handle large scale datasets, without compromising accuracy.

In summary, the known advantage of data dependency contributes to a trained model's high accuracy; whereas the two previously unknown advantages contribute to efficiency gain. These will be demonstrated in the empirical evaluations reported in Section \ref{sec_exp_results}.

\section{Isolation Kernel}
\label{sec_IK}
We provide the pertinent details of Isolation Kernel in this section. Other details can be found in \cite{ting2018IsolationKernel,IsolationKernel-AAAI2019}.

Let $D=\{{\bf x}_1,\dots,{\bf x}_n\}, {\bf x}_i \in \mathbb{R}^d$ be a dataset sampled from an unknown probability density function ${\bf x}_i \sim F$. Moreover, 
let $\mathds{H}_\psi(D)$ denote the set of
all partitionings $H$
that are admissible under the dataset $D$, 
where each $H$ covers the entire space of $\mathbb{R}^d$. Each of the $\psi$ isolating partitions $\theta[{\bf z}] \in H$ isolates one data point ${\bf z}$ from the rest of the points in a random subset $\mathcal D \subset D$, and $|\mathcal D|=\psi$, where each point in $\mathcal D$ has the equal probability of being selected from $D$.

\begin{definition} For any two points ${\bf x}, {\bf y} \in \mathbb{R}^d$,
	Isolation Kernel of ${\bf x}$ and ${\bf y}$ wrt $D$ is defined to be
	the expectation taken over the probability distribution on all partitionings $H \in \mathds{H}_\psi(D)$ that both ${\bf x}$ and ${\bf y}$  fall into the same isolating partition $\theta[{\bf z}] \in H, {\bf z} \in \mathcal{D}$:
	\begin{eqnarray}
K_\psi({\bf x},{\bf y}\ |\ D) &=&  {\mathbb E}_{\mathds{H}_\psi(D)} [\mathds{1}({\bf x},{\bf y} \in \theta[{\bf z}]\ | \ \theta[{\bf z}] \in H)] \nonumber \\
&=& {\mathbb E}_{\mathcal{D} \subset D} [\mathds{1}({\bf x},{\bf y}\in \theta[{\bf z}]\ | \ {\bf z}\in \mathcal{D})]  \nonumber
\\
&=& P({\bf x},{\bf y}\in \theta[{\bf z}]\ | \ {\bf z}\in \mathcal{D} \subset D)
		\label{eqn_kernel}
	\end{eqnarray}
where $\mathds{1}(\cdot)$ is an indicator function.
\end{definition}

In practice, Isolation Kernel $K_\psi$ is constructed using a finite number of partitionings $H_i, i=1,\dots,t$, where each $H_i$ is created using $\mathcal{D}_i \subset D$:
\begin{eqnarray}
K_\psi({\bf x},{\bf y}|D)  & = &  \frac{1}{t} \sum_{i=1}^t   \mathds{1}({\bf x},{\bf y} \in \theta\ | \ \theta \in H_i) \nonumber\\
 & = & \frac{1}{t} \sum_{i=1}^t \sum_{\theta \in H_i}   \mathds{1}({\bf x}\in \theta)\mathds{1}({\bf y}\in \theta) 
 \label{Eqn_IK}
\end{eqnarray}

$\theta$ is a shorthand for $\theta[{\bf z}]$. 

$K$ is a shorthand for $K_\psi$ hereafter.

Isolation Kernel is positive semi-definite as Eq \ref{Eqn_IK} is a quadratic form. Thus, Isolation Kernel defines a Reproducing Kernel Hilbert Space (RKHS).

\subsection{iForest implementation}
\label{sec_iForest}
Here the aim is to isolate every point in $ \mathcal D$. This is done recursively by randomly selecting an axis-parallel split to subdivide the data into two non-empty subsets until every point is isolated. Each partitioning $H$ produces $\psi$ isolating partitions $\theta$; and each partition contains a single point in  $ \mathcal D$.

The algorithm $iForest$ \cite{liu2008isolation} produces $t$ $iTrees$, each built independently using a subset $\mathcal D \subset D$, 
sampled without replacement from $D$, where $\vert \mathcal D \vert=\psi$. 

%

\subsection{aNNE Implementation}
\label{sec_aNNE}
As an alternative to using trees in its first implementation of Isolation Kernel \cite{ting2018IsolationKernel}, a nearest neighbour ensemble (aNNE) has been used instead \cite{IsolationKernel-AAAI2019}. 

Like the tree method, the nearest neighbour method also produces each $H$ model which consists of $\psi$ isolating partitions $\theta$, given a subsample of $\psi$ points. Rather than representing each isolating partition as a hyper-rectangle, it is represented as a cell in a Voronoi diagram, where the boundary between two points is the equal distance from these two points.

$H$, being a Voronoi diagram, is built by employing $\psi$ points in $\mathcal D$,
where each isolating partition or Voronoi cell $\theta \in H$ isolates one data point from the rest of the points in $\mathcal D$. The point which determines a cell is regarded as the cell centre.

Given a Voronoi diagram $H$ constructed from a sample $\mathcal{D}$ of $\psi$ points, the Voronoi cell centred at $z \in \mathcal{D}$ is:
\[ \theta[z] = \{x  \in \mathbb{R}^d \ | \  z = \argmin_{\mathsf{z} \in \mathcal{D}} \ell_p(x - \mathsf{z})\}. \]
\noindent where $\ell_p(x, y)$ is a distance function and we use $p=2$ as Euclidean distance in this paper. 

Note that the boundaries of a Voronoi diagram is derived implicitly to be equal distance between any two points in $\mathcal{D}$; and it needs not be derived explicitly for our purpose in realising Isolation Kernel.

\subsection{Kernel distributions and contour plots}
\label{sec_illustration}
\begin{figure}[!t]
 \centering
 \subfloat[Laplacian kernel]{\includegraphics[height=.16\textwidth,width=.23\textwidth]{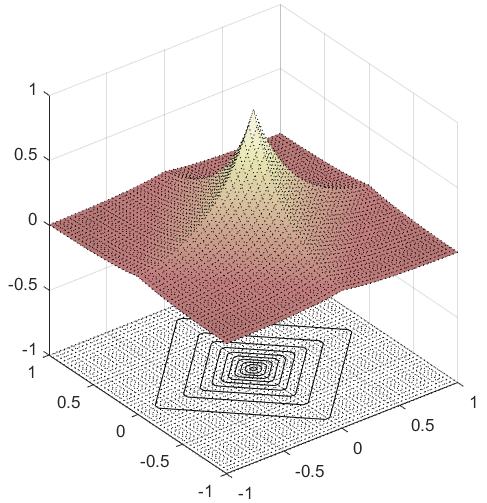}} \hspace{.01in}
 \subfloat[Isolation Kernel]{\includegraphics[height=.16\textwidth,width=.23\textwidth]{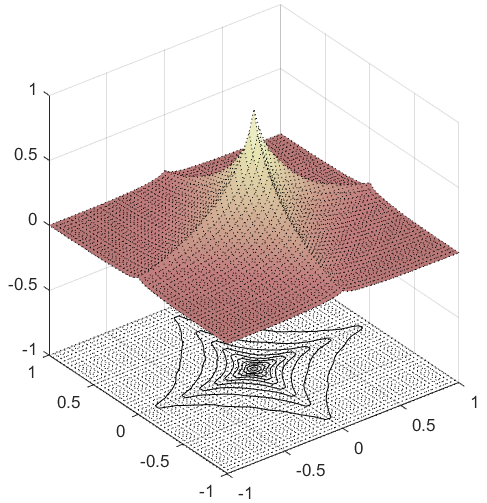}}
 \caption{Laplacian and Isolation Kernel with reference to point (0, 0) on a 2-dimensional dataset with uniform density distribution. Isolation and Laplacian use the same $\psi = 256$; and Isolation uses $t=10000$.}
 \label{fig:uniform}
\end{figure}

Figure \ref{fig:uniform} is extracted from \cite{ting2018IsolationKernel} which shows that the kernel distribution of Isolation Kernel approximates that of Laplacian kernel under uniform density distribution. A brief description of the proof is provided in the same paper.

\begin{figure}[!t]
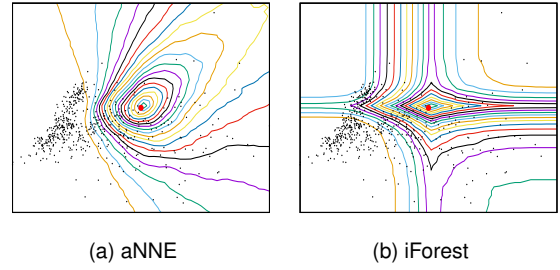

 \centering
 \subfloat[aNNE]{\includecombinedgraphics[height=.17\textwidth,width=.21\textwidth,vecfile=./aNNE_forest_psi_10_t_1000_lp_2_pdf]{./aNNE_forest_psi_10_t_1000_lp_2}}
 \subfloat[iForest]{\includecombinedgraphics[height=.17\textwidth,width=.21\textwidth,vecfile=./iForest_forest_psi_10_t_1000_lp_2_pdf]{./iForest_forest_psi_10_t_1000_lp_2}}
 \caption{Contour plots of two different implementations of Isolation Kernel on a real-world dataset. Both aNNE and iForest use $\psi=10$ and $t=1000$. Black dots are data points.}
\label{fig_contour_example}
\end{figure}

Figure \ref{fig_contour_example} shows that the contour plots of aNNE and iForest implementations of Isolation Kernel. Notice that each contour line, which denotes the same similarity to the centre (red point), is elongated along the sparse region and compressed along the dense region. In contrast, Laplacian kernel (or any data independent kernel) has the same symmetrical contour lines around the centre point, independent of data distribution (as shown in Figure \ref{fig:uniform}(a)).

The reasons why Voronoi-based implementation are better than the tree-based implementation have been provided earlier \cite{IsolationKernel-AAAI2019}; and this has led to better density-based clustering result than using the Euclidean distance measure. 

\section{Learning with exact feature map}
\label{sec_learning_IK}

This section presents our four conceptual contributions.
Section \ref{sec_directfeaturemap} presents the feature map of Isolation Kernel. Section \ref{sec_learn_kernel} describes the theoretical underpinning of efficient learning with Isolation Kernel. How Isolation Kernel enables the use of $_{primal}f$ in solving the primal optimisation problem, and its efficient dot product implementations are provided in the following two subsections.

\subsection{Exact feature map of Isolation Kernel}
\label{sec_directfeaturemap}
Viewing each isolating partition $\theta$ as a feature, the $i$-component of the feature space due to $H_i$
can be derived using the mapping 
$\Phi_i:\mathcal{X} \mapsto \mathbb{I}^{\psi}$ (where $\mathbb{I}$ is a binary domain); and $\forall {\bf x}, {\bf y} \in \mathbb{R}^d,\ {\bf x}, {\bf y} \mapsto K^i({\bf x}, {\bf y})$  can be constructed  using a partitioning $H_i$ as follows:

Let $\Phi_i({\bf x})=[\mathds{1}({\bf x} \in \theta_1), \dots, \mathds{1}({\bf x} \in \theta_\psi)]^\top$  be a vector of $\psi$ features of $\{0,1\}$ indicating the only isolating partition in which ${\bf x}$ falls, out of the $\psi$ isolating partitions $\theta_1,\dots,\theta_\psi$, where $H_i=\{\theta_j|\ j=1,...,\psi\}$.

The inner summation of Equation (\ref{Eqn_IK}) of Isolation Kernel can then be re-expressed in terms of $\Phi_i$ as follows:
\[\sum_{\theta \in H_i}   \mathds{1}({\bf x}\in \theta)\mathds{1}({\bf y}\in \theta) =\Phi_i({\bf x})^\top\Phi_i({\bf y}) = K^i({\bf x},{\bf y})\]

Because $K^i$ is in a quadratic form, it is a PSD (positive semi-definite).
The sum of PSD, $K = \frac{1}{t} \sum_{i=1}^t K^i$, is also PSD. Therefore, $K$ is a valid kernel.

An exact simple representation of Isolation Kernel can be derived by concatenating $t$ samples of $\mathcal{D} \subset D$. Let $\Phi(\cdot)$ be a vector of $t\psi$ binary features. Then, Isolation Kernel represented using these features can be expressed as:
\[
K({\bf x},{\bf y}) =  \frac{1}{t} \left<\Phi({\bf x}),\Phi({\bf y})\right>  
\]

\begin{definition}
	\label{def:featureMap}
	\textbf{Feature map of Isolation Kernel.}
	For point ${\bf x} \in \mathbb{R}^d$, the feature mapping $\Phi: {\bf x}\rightarrow \mathbb \{0,1\}^{t\times \psi}$ of $K$ is a vector that represents the partitions in all the partitioning $H_i\in \mathds{H}_\psi(D)$, $i=1,\dots,t$; where ${\bf x}$ falls into only one of the $\psi$ partitions in each partitioning $H_i$.
\end{definition}

Let $\mathbbmtt{1}$ be a shorthand of $\Phi_i(x)$ such that $\Phi_{ij}(x)=1$ and $\Phi_{ik}(x)=0, \forall k \neq j$ for any $j \in [1,\psi]$.

$\Phi$ is sparse and has the following geometrical interpretation:
 Every mapped point is denoted as  $\left[\mathbbmtt{1}, \dots, \mathbbmtt{1}  \right]$ but it is not a single point in RKHS, but points which have $\parallel {\Phi}(x) \parallel\ = \sqrt{t}$ and $\Phi_i(x) = \mathbbmtt{1}$ for all $i \in [1,t]$; and
  $K(x,x|D)=1$  for all $x \in \mathbb{R}^d$.

Parameters $t$ and $\psi$ can be controlled by a user. Each setting of $t$ and $\psi$ yields a feature map.

\begin{table*}[!t]
 \caption{Feature map construction comparison: Nystr\"{o}m versus Isolation Kernel.}
 \label{tab:feature_map_construction}
 \begin{tabular}{lp{8.5cm}|lp{8.2cm}}
  \toprule
& Nsytr\"{o}m (approximate feature map from a chosen kernel $K(\cdot,\cdot)$) & & Isolation Kernel ($\phi({\bf x})$) \\
  \midrule
1. & Sample $\{{\bf x}_i\ |\ i=1,\dots,b\}$ from $D$ 
to construct kernel matrix $\mathbf{K}$ & 1. & Sample $\psi$ points from $D$, $t$ times, to construct $t$ partitionings $H_i \in \mathbb{H}_\psi(D)$; and each $H_i$ has $\psi$ partitions.\\
2. & $[\mathbf{V}_r, \mathbf{A}_r] = eigens(\mathbf{K},r)$, \hspace{5cm} \linebreak where $\mathbf{V}_r$ and $\mathbf{A}_r$ are eigenvectors and eigenvalues of $\mathbf{K}$. & 2. & \multirow{2}{8cm}{${\bf x} \in {\mathbb R}^d \rightarrow \phi({\bf x}) \in \mathbb{Z}^t$, where each integer attribute has values: $1,\dots,\psi$; and each integer is an index to a partition $\theta \in H_i$. \linebreak The $t$ attributes represent the partitionings $H_i, i=1,\dots,t$.\linebreak Convert ${\bf x} \in D$ to $\phi({\bf x}) \in \acute{D}$: ${\bf x}$ is parsed over the $t$ partitionings.}\\
3. & ${\bf x} \in {\mathbb R}^d \rightarrow \acute{{\bf x}} \in \mathbb{R}^r$:\hspace{5.2cm} \linebreak Convert ${\bf x} \in D$ to $\acute{{\bf x}} \in \acute{D}$:\ $\acute{{\bf x}} = \mathbf{A}_r^{-0.5} \mathbf{V}_r^{\top} (K({\bf x}, {\bf x}_1), \dots, K({\bf x}, {\bf x}_b))^{\top}$\\
  \midrule
 \multicolumn{4}{c}{Perform learning with feature map on $\acute{D}$}  \\
  \bottomrule
 \end{tabular}
\end{table*}

\subsection{Efficient Learning with Isolation Kernel}
\label{sec_learn_kernel}

This subsection describes the theoretical underpinning of efficient learning with Isolation Kernel.

In a binary class learning problem of a given training set $D=\{({\bf x}_1,c_i),\dots,({\bf x}_n,c_n)\}$, where points ${\bf x}_i \in \mathbb{R}^d$ and class labels  $c_i \in \mathcal{C}=\{+1, -1\}$,
the goal of SVM is to learn a kernel prediction function $f$ by solving the following optimisation problem \cite{LearningWithKernelsBook2001}:
\[
\min_{f \in \mathcal{H}_D} \frac{\lambda}{2} \parallel f \parallel_{\mathcal{H_K}}^2 + \frac{1}{n} \sum_{i=1}^n L(f({\bf x}_i); c_i)
\]

\noindent
where $\mathcal{H}_D = span(K(\cdot, {\bf x}_1), \dots, K(\cdot, {\bf x}_n))$ is span over all points in the training set $D$; $L(f({\bf x}); c)$ is a convex loss function wrt the prediction of ${\bf x}$; and $\mathcal{H_K}$ is the Reproducing Kernel Hilbert Space endowed with a kernel.

The computational cost of this kernel learning is high because the search space over $\mathcal{H}_D$ is large for large $n$.

In contrast, with Isolation Kernel, $\mathcal{H}_D$ is replaced with a smaller set $\mathcal{H}_T = span(\Phi_1(\cdot), \Phi_2(\cdot), \dots, \Phi_t(\cdot)) \subset [0,1]^{t\psi}$. 

When $|\cup_{i=1}^t \mathcal{D}_i| \ll |D|$ which leads to $\mathcal{H}_T \ll \mathcal{H}_D$, learning with Isolation Kernel is expected to be faster than learning with commonly used data independent kernels such as Gaussian and Laplacian kernels.

The following subsections provide the implementations---due to the use of Isolation Kernel---which enable the significant efficiency gain without compromising predictive accuracy for online kernel learning.

\subsection{Using $_{primal}f$ instead of $_{dual}f$}
\label{sec_testing}

The prediction function employed follows the respective functional form of either the dual or the primal optimisation problem in which one is solving. 

When existing kernels such as Gaussian and Laplacian kernels are used, because they have infinite number of features,  
the dual optimisation problem and $_{dual}f({\bf x}) = \sum_{i=1}^s \alpha_i c_i K({\bf x}_i,{\bf x})$ must be used (unless an approximate feature map is derived).

As Isolation Kernel has a finite-dimensional feature map,  this facilitates the use of prediction function 
 $_{primal}f({\bf x}) = \left<{\bf w}, \Phi({\bf x})\right>$; thus 
solving the primal optimisation problem is a natural choice.

The evaluation of $_{primal}f$ is faster than that of $_{dual}f$, when the number of support vectors ($s$) times the number of attributes of ${\bf x}$  ($d$) is more than the effective number of features of $\Phi({\bf x})$, i.e., $sd > t$ (see the reason why $t$ is the effective number of feature of the $\Phi({\bf x})$ in next subsection). Its use yields a significant speedup when the domain is high dimensional and/or in an online setting where the points can potentially be infinite. The online setting necessitates the need to have a kernel learning system which can deal with potentially infinite number of support vectors. The procedure of such a kernel learning system using Isolation Kernel is described in Section \ref{sec_applications}.

\subsection{Efficient dot product in $_{primal}f$}
\label{sec_efficient_dot_product}
The use of Isolation Kernel facilitates an efficient dot product $\left<{\bf w}, \Phi({\bf x})\right>$ in $_{primal}f$. Recall that, $\forall {\bf x} \in \mathbb{R}^d$, $\Phi_i({\bf x})$ has exactly one feature having value=1 in a vector of $\psi$ binary features (stated in Section \ref{sec_directfeaturemap}).
Thus, $\left<{\bf w}, \Phi({\bf x})\right>$ can be computed with a summation of $t$ number of $w_{ij}$ (rather than the naive dot product, computing $t\psi$  products $w_{ij} \times \Phi_{ij}({\bf x})$):
\[ \left<{\bf w}, \Phi({\bf x})\right> = \sum_{i=1}^t \sum_{j=1}^\psi w_{ij} \times \Phi_{ij}({\bf x}) = \sum_{i=1, j=\phi_{i}({\bf x}) \in \mathbb{Z}}^t  w_{ij} \]
\noindent
where $\Phi_{ij}({\bf x})$ denotes the value of binary feature $j$ of $\Phi_{i}({\bf x})$; and $\phi_{i}({\bf x})=j$ serves as an index to the $j$-th element of $\Phi_i({\bf x})$ indicating ${\bf x} \in \theta_j$.

In summary, $\left<{\bf w}, \Phi({\bf x})\right>$ can be computed more efficiently using $\phi$ as an indexing scheme.

Note that this efficient dot product is independent of $\psi$. For large $\psi$, this dot product could result in orders of magnitude faster than using the naive dot product (see Figure~\ref{fig_varying_psi} in Section \ref{sec_effect_efficient_dot_product} later).

The indexing scheme $\phi$ of the feature map  of Isolation Kernel is constructed in two steps as shown in Table \ref{tab:feature_map_construction} that convert ${\bf x} \in {\mathbb R}^d \rightarrow  \phi({\bf x})  \in \mathbb{Z}^t$. The steps taken by the Nystr\"{o}m method \cite{Nystrom_NIPS2000,LargeScaleOnlineKernelLearning} to construct an approximate feature map is also shown for comparison in the same table. 

The computational cost of the mapping from ${\bf x}$ to either $\Phi({\bf x})$ or $\phi({\bf x})$ is linear to $t\psi$. But this mapping needs to be done only once for each point. That is, every point needs to examine each partitioning $H$ only once to determine the partition $\theta \in H$ into which the point falls.

\section{Applications to Kernel learning that uses Online Gradient Descent and support vector machines}
\label{sec_applications}

Online kernel learning aims to build an efficient and scalable kernel-based predictive model incrementally from a sequence of potentially infinite data points. One of the early methods is \cite{OnlineLearningwithKernels2001}. One key challenge of online kernel learning is managing a growing number of support vectors, as every misclassified point is typically added to the set of support vectors. 

One recent implementation of online kernel learning is called OGD \cite{LargeScaleOnlineKernelLearning} which employs $_{dual}f$:
\[
f({\bf x}) = \sum_{i=1}^{s} \alpha_i c_i K({\bf x}_i, {\bf x})
\]
If $\triangledown L(f({\bf x}); c) \ne 0$ (incorrect prediction) then add ${\bf x}$ to the set of support vectors with $\alpha = -\eta \triangledown L(f({\bf x}); c)$, where $\eta$ is the learning rate.

Without setting a budget, the number of support vectors ($s$) usually increases linearly with the number of points observed. Therefore, the testing time becomes increasingly slower as the number of points observed increases.

Here we show the benefits of Isolation Kernel will bring to online kernel learning: 
Its use improves both the time and space complexities of OGD significantly from $O(sd)$ to $O(t \psi)$ 
for every prediction while {\em allowing $s$ to be infinite---eliminating the need to have a budget for support vectors}. This is because  $t \psi$ is constant while $s$ grows as more points are observed.

This is done on exactly the same OGD implementation. The only change required in the procedure is that the function $f$ is evaluated based on its feature map $\Phi$ of Isolation Kernel as follows: 
\begin{eqnarray}
f({\bf x}) 
 =  \sum_{i=1}^{s} \alpha_i c_i \left< \Phi({\bf x}_i), \Phi({\bf x}) \right> 
 =  \left<{\bf w}, \Phi({\bf x})\right>,\nonumber
\end{eqnarray}

\noindent
where ${\bf w} = \sum_{i=1}^{s} \alpha_i c_i \Phi({\bf x}_i)$.

During training, $s$ is the number of support vectors at the time an evaluation of the prediction function is required. For every addition of a new support vector ${\bf x}$ during the training process, the weight vector ${\bf w} = {\bf w} + \alpha c \Phi({\bf x})$ is updated incrementally while $s$ increments. At the end of the training process, the final ${\bf w}$ is ready to be used with $_{primal}f({\bf x}) = \left<{\bf w}, \Phi({\bf x})\right>$ to evaluate every test point ${\bf x}$. 

Although the above expressions are in terms of $\Phi$, the computation is conducted more efficiently using $\phi$, effectively as an indexing scheme for $\Phi$, as described in Section \ref{sec_efficient_dot_product}, for $\left<{\bf w}, \Phi({\bf x})\right>$ as well as $\sum_{i=1}^{s} \alpha_i c_i \Phi({\bf x}_i)$.

We named the OGD implementation which employs Isolation Kernel and $_{primal}f$ as IK-OGD. The algorithms of OGD (as implemented by  \cite{LargeScaleOnlineKernelLearning}) and IK-OGD are shown as Algorithms \ref{alg_OGD} and \ref{alg_IK-OGD}, respectively.

    \begin{algorithm}[!htbp]
        \caption{$OGD(\eta, K)$}
        \label{alg_OGD}
        \begin{algorithmic}[1]  
            \Require $\eta$ - learning rate; $K$ - Kernel.
        \State Initialize set of support vectors $S = \emptyset$;
    
        \While {There is a new point $\bf x$};

        \State $f({\bf x}) = \sum_{i=1}^{|S|} \alpha_i c_i K({\bf x}_i, {\bf x})$;

        \If {$\triangledown L(f({\bf x}); c) \ne 0$ (incorrect prediction)}
        \State Add ${\bf x}$ to $S$ with $\alpha = -\eta \triangledown L(f({\bf x}); c)$;
        \EndIf
    \EndWhile
         \end{algorithmic}
    \end{algorithm}

    \vspace{-1mm}
    \begin{algorithm}[!htbp]
        \caption{$IK$-$OGD(\eta, \Phi)$}
        \label{alg_IK-OGD}
        \begin{algorithmic}[1]  
            \Require $\eta$ - learning rate; $\Phi$ - feature mapping of IK.
        \State Initialize ${\bf w}$ to ${\bf 0}$;
  
        \While {There is a new point $\bf x$};

        \State $f({\bf x}) = \left<{\bf w}, \Phi({\bf x})\right>$;

        \If {$\triangledown L(f({\bf x}); c) \ne 0$ (incorrect prediction)}
        \State $\alpha = -\eta \triangledown L(f({\bf x}); c)$;
        \State ${\bf w} = {\bf w} + \alpha c \Phi({\bf x})$;
        \EndIf
    \EndWhile
         \end{algorithmic}
    \end{algorithm}

To apply Isolation Kernel to support vector machines, we only need to use the algorithm which solves the primal optimisation problem such as LIBLINEAR \cite{LIBLINEAR} after converting the data using the feature map of Isolation Kernel.

When using a kernel having a feature map with intractable dimensionality, a similar efficiency gain can be achieved by employing a kernel functional approximation method to get an approximate finite-dimensional feature map. This comes with a cost of reduced accuracy because of the approximation.

\section{Experimental settings}
\label{sec_experiments}
We design experiments to evaluate the impact of Isolation Kernel on Online Kernel Learning. We use the implementations of the kernelised online gradient descent (OGD) and Nystr\"{o}m online gradient descent (NOGD)\footnote{Codes available at \url{http://lsokl.stevenhoi.org/}.}. 
The kernelised online gradient descent \cite{OnlineLearningwithKernels2001} or OGD solves the dual optimisation problem; whereas IK-OGD solves the primal optimisation problem, so as NOGD \cite{LargeScaleOnlineKernelLearning}. 
We also compare with a recent online method that employs multi-kernel learning and random fourier features, called AdaRaker \cite{OnlineMKL-2018}.

Laplacian kernel is used as a base-line kernel because Isolation Kernel approximates Laplacian kernel under uniform density distribution \footnote{As pointed in \cite{ting2018IsolationKernel}, Laplacian kernel can be expressed as $L_\psi({\bf x},{\bf y}) = \exp(-\lambda \sum^d_{\jmath=1} |x_\jmath - y_\jmath|)
= \psi^{-\frac{1}{d} \sum^d_{\jmath=1} |x_\jmath - y_\jmath|}$, where $\lambda = \frac{\log(\psi)}{d}$. Laplacian kernel has been shown to be competitive to Gaussian kernel in SVM in a recent study \cite{ting2018IsolationKernel}.}. As a result, Isolation Kernel and Laplacian kernel can be expressed using the same `sharpness' parameter $\psi$.

Two existing implementations of Isolation Kernel are used: (i) Isolation Forest \cite{liu2008isolation}, as described in \cite{ting2018IsolationKernel}; and (ii) aNNE, a nearest neighbour ensemble that partitions the data space into Voronoi diagram, as described in \cite{IsolationKernel-AAAI2019}. We refer IK-OGD to the iForest implementation. When a distinction is required, we denote IK$_i$-OGD as the iForest implementation; and IK$_a$-OGD the aNNE implementation.

All OGD related algorithms used the hinge loss function and the same learning rate $\eta=0.5$, as used in \cite{LargeScaleOnlineKernelLearning}. The only parameter search required for these algorithms is the kernel parameter. The search range in the experiments is listed in Table~\ref{tbl:param}. The parameter is selected via 5-fold cross-validation on the training set.

The default settings for NOGD \cite{LargeScaleOnlineKernelLearning} are: the Nystr\"{o}m method uses the Eigenvalue-Decomposition; and sampling size or budget\footnote{Note that this parameter is called budget in \cite{LargeScaleOnlineKernelLearning}; but this is different from the budget used to limit the number of support vectors, mentioned in Section \ref{sec_challenges}.} $b=100$; and the matrix rank is set to $r=0.2b$. The default parameter used to create Isolation Kernel is set to $t=100$.

AdaRaker (https://github.com/yanningshen/AdaRaker) employs sixteen Gaussian kernels and the specified bandwidths $\sigma$ for these kernels are listed in Table~\ref{tbl:param}. (The default three kernels in the code gave worse accuracy than that reported in the next section). In addition, AdaRaker uses 50 orthogonal random features (equivalent to $b=100$ for the Nystr\"{o}m method) and $\eta=0.5$ as default. The search range of $\lambda$ through 5-fold cross-validation is given in Table~\ref{tbl:param}.

 \begin{table}[!t]
 	\centering
 	\caption{Search ranges of parameters.} \label{tbl:param}
 	\begin{tabular}{cl}\hline
 Kernel/Algorithm	 & Search range \\\hline
Laplacian  &  \multirow{2}{*}{$\psi \in \{ 2^m\ |\ m=2,3,\dots,12\}$}  \\ \cline{1-1}
 		Isolation &    \\\hline
 \multirow{2}{*}{AdaRaker} & $\lambda \in \{10^m,\frac{10^m}{2}\ |\ m=-2,-3,-4,-5\}$\\
  &   $\sigma \in \{ 2^m\ |\ m=-10,\dots,4,5 \}$ \\\hline
 	\end{tabular}
 \end{table}

Eleven datasets from \url{www.csie.ntu.edu.tw/\~cjlin/libsvmtools/datasets/} are used in the experiments.
The properties of these datasets are shown in Table~\ref{tbl:data_properties}. The datasets are selected in order to have diverse data properties: data sizes (20,000 to 2,400,000) and dimensions (22 to more than 3.2 million).
Because the OGD and NOGD versions of the implementation we used work on two-class problems only, three multi-class datasets have been converted to two-class datasets of approximately equal class distribution\footnote{The two-class conversions from the original class labels were done for three multi-class datasets: mnist: $\{3, 4, 6, 7, 9\}$ and $\{0, 1, 2, 5, 8\}$. smallNORB: $\{1,4\}$ and $\{0, 2, 3\}$. cifar-10: $\{ 0, 2, 3, 4, 5\}$ and $\{1, 6, 7, 8, 9\}$.}.

Four experiments are conducted: (a) in online setting, (b) in batch setting, (c) examine the runtime in GPU, and (d) an investigation using SVM. The CPU experiments ran on a Linux CPU machine: AMD 16-core CPU with each core running at 1.8 GHz and 64 GB RAM. The GPU experiments ran on a machine having 
GPU: 2 x GTX 1080 Ti with 3584 (1.6 GHz) CUDA cores \& 12GB graphic memory; and CPU: i9-7900X 3.30GHz processor (20 cores), 64GB RAM.

 The results are presented in four subsections of Section~\ref{sec_exp_results}.

\begin{table}[!t]
 \centering
 \caption{Properties of the datasets used in the experiments. $nnz\% = \#nonzero\_values / ((\#train+\#test) \times \#dimensions) \times 100$. Datasets with $nnz\% < 1\%$ are regarded as sparse datasets; otherwise, they are dense datasets.
 \label{tbl:data_properties} }
 \begin{tabular}{c|rrrr}\toprule
			& \#train     & \#test & \#dimensions & nnz\% \\
	\midrule
    url & 30,000 & 2,366,130 & 3,231,961 & 0.0036 \\
    news20.binary & 15,997 & 3,999 & 1,355,191 & 0.03 \\
    rcv1.binary & 20,242 & 677,399 & 47,236 & 0.16 \\
    real-sim & 57,848 & 14,461 & 20,958 & 0.24 \\
    smallNORB & 24,300 & 24,300 & 18,432 & 100.0 \\
    cifar-10 & 50,000 & 10,000 & 3,072 & 99.8 \\
    epsilon & 400,000 & 100,000 & 2,000 & 100.0 \\
    mnist & 60,000 & 10,000 & 780 & 19.3 \\
    a9a & 32,561 & 16,281 & 123 & 11.3 \\
    covertype & 464,810 & 116,202 & 54 & 22.1 \\
    ijcnn1 & 49,990 & 91,701 & 22 & 59.1 \\
  \bottomrule
 \end{tabular}
\end{table}

In online setting, we simulate an online setting  using each of the four largest datasets over half a million points (after combining their given training and testing sets) as follows. 
Given a dataset, it is first shuffled. Then, the initial training set has data size as the training set size shown in Table \ref{tbl:data_properties}; and it is used to determine the best parameter based on 5-fold cross-validation before training the first model.  The online stream is assumed to arrive sequentially in blocks of 1000 points. Each block is assumed to have no class labels initially: In testing mode, the latest trained model is used to make a prediction for every point in the block. After testing, class labels are made available: The block is in training mode and the model is updated\footnote{This simulation is more realistic than the previous online experiments which assume that class label of each point is available immediately after a prediction is made to enable model update \cite{LargeScaleOnlineKernelLearning}. In practice, the algorithm can be made to be in the training mode whenever class labels are available, either partially or the entire block.}. The above testing and training modes are repeated for each current block in the online stream until the data run out. The test accuracy up to the current block is reported along the data stream. 

In batch setting, we report the result of a single trial of train-and-test for each dataset which consists of separate training set and testing set. The assessments are in terms of predictive accuracy and the total runtime of training and testing. Since AdaRaker has problem dealing with large datasets, it is used in the batch setting only.

In online setting, Isolation Kernel and $\phi$ for IK-OGD are established using the initial training set only. Once established, the kernel and $\phi$ are fixed for the rest of the data stream. This applies to the $b$ points selected for NOGD as well. In the batch setting, the given training set is used for these purposes.

\section{Empirical Results}
\label{sec_exp_results}
\subsection{Results in online setting}

\begin{figure}[!t]
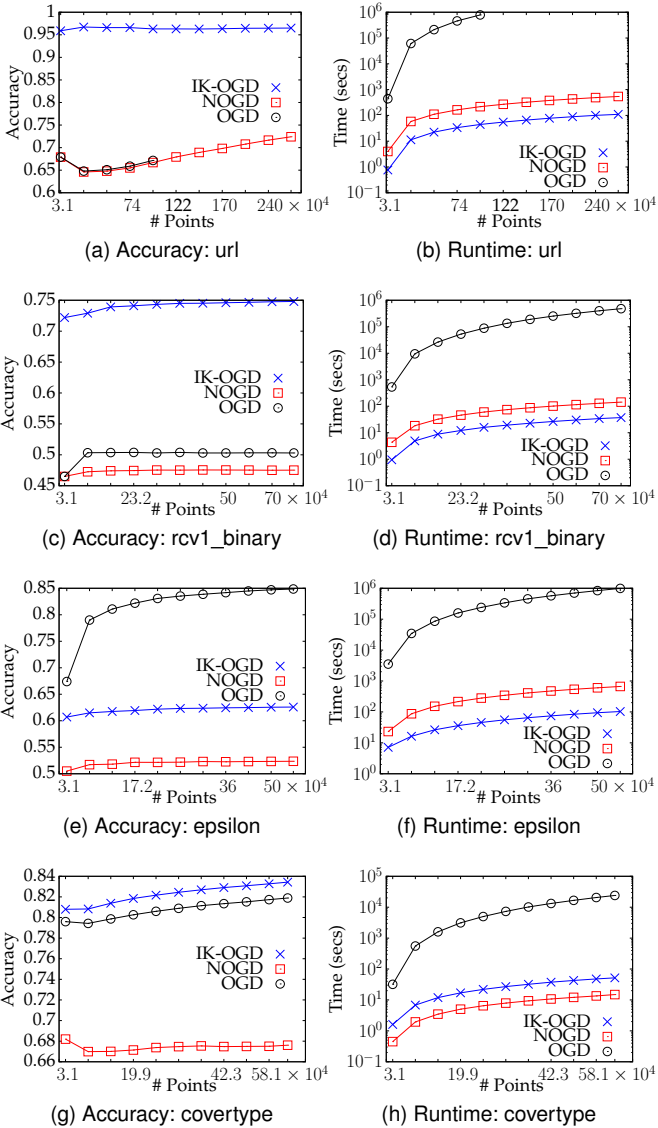

 \centering
 
 \subfloat[Accuracy: url]{\includecombinedgraphics[scale=0.34,vecfile=./url_accuracy_v2_pdf]{./url_accuracy_v2}}
 \subfloat[Runtime: url]{\includecombinedgraphics[scale=0.34,vecfile=./url_timing_v2_pdf]{./url_timing_v2}}\\

 \subfloat[Accuracy: rcv1\_binary]{\includecombinedgraphics[scale=0.34,vecfile=./rcv1_accuracy_v2_pdf]{./rcv1_accuracy_v2}}
 \subfloat[Runtime: rcv1\_binary]{\includecombinedgraphics[scale=0.34,vecfile=./rcv1_timing_v2_pdf]{./rcv1_timing_v2}}\\

 \subfloat[Accuracy: epsilon]{\includecombinedgraphics[scale=0.34,vecfile=./epsilon_accuracy_v2_pdf]{./epsilon_accuracy_v2}}
 \subfloat[Runtime: epsilon]{\includecombinedgraphics[scale=0.34,vecfile=./epsilon_timing_v2_pdf]{./epsilon_timing_v2}}\\

 \subfloat[Accuracy: covertype]{\includecombinedgraphics[scale=0.34,vecfile=./covtype_accuracy_v2_pdf]{./covtype_accuracy_v2}}
 \subfloat[Runtime: covertype]{\includecombinedgraphics[scale=0.34,vecfile=./covtype_timing_v2_pdf]{./covtype_timing_v2}}\\

 \caption{Results in online setting in terms of accuracy and runtime (which were what each algorithm could complete within one week.) }
 \label{fig_online}
\end{figure}

Figure \ref{fig_online} shows that, in terms of accuracy, IK-OGD has higher accuracy than OGD and NOGD on four datasets, except that OGD has better accuracy on epsilon only\footnote{Note that the first points in the accuracy plots can swing wildly because it is the accuracy 
on the first data block.}. Notice that, as more points are observed, OGD and IK-OGD have more rooms for accuracy improvement than NOGD because the former two have no budget and the latter has a limited budget. We will examine the extent to which increasing the budget and $t$ improve the accuracies of NOGD and IK-OGD, respectively, in Section~\ref{sec_batch_setting}.

In terms of runtime, IK-OGD runs faster than both OGD and NOGD on high dimensional datasets (url, rcv1.binary and epsilon); and it is only slower than NOGD in the low dimensional covertype dataset.
Notice that the gap in runtime between IK-OGD and NOGD stays the same over the period because the time spent on $_{primal}f$ is the same. In contrast, the gap between OGD and IK-OGD increases over time because the time spent on $_{dual}f$ used by OGD increases as the number of support vectors increases over time. The runtimes of IK-OGD and NOGD are in the same order; but IK-OGD is 2 to 4 orders magnitude faster than OGD; and on url, ODG could only complete the first five points in Figures~\ref{fig_online}(a) \& \ref{fig_online}(b) after more than one week.

NOGD maintains fast execution by limiting the number of support vectors while using $_{primal}f$. The use of Laplacian kernel (or any other kernel) which has infinite or large number of features necessitates the use of a feature map approximation method.  Despite all these measures for efficiency gain in NOGD, IK-OGD without budget still ran faster than NOGD with budget ($b=100$) on the three high-dimensional datasets!
The efficiency gain in NOGD is a trade-off with accuracy---both the feature map approximation and the limit on the number of support vectors reduce the accuracy. 

The use of Isolation Kernel provides a cleaner and simpler utilisation of $_{primal}f$ in online setting than the kernel functional approximation approach (in which NOGD is a good representative method). As a result, IK-OGD achieves the efficiency gain without compromising the accuracy because an exact rather than an approximate feature map is used.

The next two subsections provide empirical evidence of efficiency gains in IK-OGD, described in Sections \ref{sec_testing} and \ref{sec_efficient_dot_product}.

\subsubsection{The effect of $_{primal}f$ or $_{dual}f$ on IK-OGD}
To demonstrate the impact of the type of prediction function used in IK-OGD (stated in Section \ref{sec_testing}), we create a version which  employs $_{dual}f$ named IK-OGD(dual) to compare with IK-OGD which employs $_{primal}f$.

The proportions of time spent on the two prediction functions out of the total runtimes are given as follows: IK-OGD took 
2.3\% and 0.77\%
on rcv1.binary and epsilon, respectively. In contrast, IK-OGD(dual) took 99.9\% and 99.8\%, respectively. This shows that $_{primal}f$ has reduced the time spent on the prediction function from almost the total runtime to a tiny fraction of the total runtime!

The total runtimes of IK-OGD versus IK-OGD(dual) are
37 seconds versus 280,656 seconds on rcv1.binary; and  103 seconds versus 235,966 seconds on epsilon. %
In other words,  it also reduced the total runtime significantly by 3 to 4 orders of magnitude. The difference in runtimes enlarges as more points are observed because the number of support vectors increases which affects IK-OGD(dual) only. The number of support vectors used at the end of the data stream is: 349,009 for rcv1.binary; and 349,481 for epsilon. 

\begin{figure}[!t]
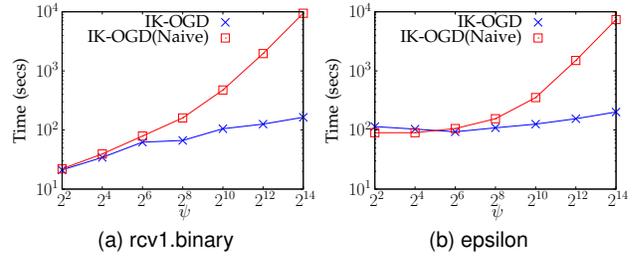

 \subfloat[rcv1.binary]{\includecombinedgraphics[scale=0.325,vecfile=./rcv1_iF_fast_slow_v2_pdf]{./rcv1_iF_fast_slow_v2}}
 \subfloat[epsilon]{\includecombinedgraphics[scale=0.325,vecfile=./epsilon_iF_fast_slow_v2_pdf]{./epsilon_iF_fast_slow_v2}}
 \caption{Runtime comparison: IK-OGD vs IK-OGD(naive) with increasing $\psi$ ($t=100)$ in online setting. }
 \label{fig_varying_psi}
\end{figure}

\begin{table*}[!t]
 \centering
 \caption{Comparing IK-OGD with OGD, NOGD and AdaRaker: Accuracy, total runtime of training and testing in seconds. ODG and NOGD use Laplacian kernel; IK$_i$-OGD uses Isolation Kernel implemented with iForest; and IK$_a$-OGD uses Isolation Kernel implemented with aNNE. AdaRaker employs 16 Gaussian kernels. The best and the worst accuracies on each dataset are boldfaced and underlined, respectively. The same applies to runtime. \newline \textcolor{green}{-ME-} denotes memory errors.
 \label{tbl:main_results}}
 \vspace{-9pt}
 \begin{tabular}{lrrrrr||rrrrrr}
  \toprule
  & \multicolumn{5}{c||}{Accuracy} &  \multicolumn{6}{c}{Runtime (CPU seconds)} \\
 & OGD & IK$_i$-OGD & IK$_a$-OGD & NOGD & AdaRaker & {OGD} & {IK$_i$-OGD} & \multicolumn{2}{c}{IK$_a$-OGD} & {NOGD} & AdaRaker \\
  \midrule
url  & .\underline{67} & .{\bf 96} & .{\bf 96} & .\underline{67} & --- & \underline{65,319} & {\bf 62} & \multicolumn{2}{c}{\cellcolor{lightgray}} & 303 & \textcolor{green}{-ME-} \\
news20.binary  &	.\underline{50}	&	.57	& 	.{\bf 89}	&	.\underline{50} & --- & \underline{915} &	{\bf 1}	&  \multicolumn{2}{c}{\cellcolor{lightgray} \multirow{10}{1.4cm}{\textcolor{gray}{See Section~\ref{sec_GPUruntime} \& Table~\ref{tbl:GPU_results}} } } 	&	11 & \textcolor{green}{-ME-} \\
rcv1.binary 	&	.\underline {48}	&	.73	&	.{\bf 96}	&	.\underline{48} & --- &	\underline{10,499}	&	{\bf 22}	& \multicolumn{2}{c}{\cellcolor{lightgray}} 	& 114 & \textcolor{green}{-ME-} \\
real-sim 	&	.73	&	.83	&	.{\bf 96}	&	.\underline{69} & --- &	\underline{1,468}	&	{\bf 2}	& \multicolumn{2}{c}{\cellcolor{lightgray}}	& 6 & \textcolor{green}{-ME-}	\\
smallNORB 	&	.{\bf 93}	&	.78	&	 .88	&	.\underline{51} & --- &	\underline{64,183}	&	{\bf 73}	& &		& 	353 & \textcolor{blue}{$>$ 1 week} \\
cifar-10 	&	.69	&	.72	&	.{\bf 73} 	&	.\underline{50}	& .54 &	\underline{20,260}	& {\bf 15} & & & 69 & 5,661\\
epsilon 	&	.{\bf 88}	&	.65	&	.71 	&	.\underline{57} & --- &	\underline{496,065}	&	{\bf 106} & 	 	&	 	 &	 430 & \textcolor{blue}{$>$ 1 week} \\
mnist 	&	.97	&	.95	& .{\bf 98} &	.85	& .\underline{80} & 659	& {\bf 4}	& & & 12 & \underline{1,453} \\
a9a 	&	.{\bf 84}	&	.{\bf 84}	&	.{\bf 84}	&	.{\bf 84} & .\underline{79}  &	95	&	3	& 	\multicolumn{2}{c}{\cellcolor{lightgray}}	 &	{\bf 2} & \underline{308} \\
covertype 	&	.76	&	.86	& .{\bf 92} &	.\underline{70}	& .\underline{70} &	\underline{20,863}	&	25	& 	 \multicolumn{2}{c}{\cellcolor{lightgray}}	&	{\bf 10} & 3,740 \\
ijcnn1 	&	.94	&	.95	&	.{\bf 97}	&	.93	& .\underline{90}  &	76	&	8	& \multicolumn{2}{c}{\cellcolor{lightgray}} &	{\bf 2} & \underline{576} \\
   \bottomrule
 \end{tabular}
\end{table*}

\subsubsection{The effect of efficient dot product on IK-OGD}
\label{sec_effect_efficient_dot_product}
Here we show the effect of the efficient dot product, described in Section~\ref{sec_efficient_dot_product}. The implementation which computes the summation of $t\psi$ products is named IK-OGD(naive). It is compared with IK-OGD with the efficient implementation.
As the impact on runtimes varies with $\psi$, the experiment is conducted with increasing $\psi$.

Figure \ref{fig_varying_psi} shows that the runtime difference between IK-OGD and IK-OGD(naive) enlarges as $\psi$ increases; and IK-OGD(naive) was close to two orders of magnitude slower than IK-OGD at $\psi=16384$ on both datasets. Note that the efficient dot product in IK-OGD is independent of $\psi$. IK-OGD's runtime depends on $\psi$ only in the process of mapping ${\bf x}$ to $\phi({\bf x})$ (recall the mapping stated in  Table \ref{tab:feature_map_construction}).

\subsection{Results in batch setting}
\label{sec_batch_setting}
Observations from the results shown in Table \ref{tbl:main_results}  are:

In terms predictive accuracy:
\begin{itemize}
\item  IK$_i$-OGD performs better than OGD on six datasets; it has equal or approximately equal accuracy on mnist, a9a and ijcnn1.  This outcome is purely due to the kernel employed---Isolation Kernel approximates Laplacian kernel under uniform density distribution; and it adapts to density structure of the given dataset \cite{ting2018IsolationKernel}.  This relative result between Isolation Kernel and Laplacian Kernel on OGD is consistent with the previous relative result on SVM \cite{ting2018IsolationKernel}.
The only two datasets on which IK$_i$-OGD performs significantly worse than OGD  are smallNORB and epsilon. We will see in Section \ref{sec_effects} that the gap can be significantly reduced by increasing $t$, without a significant runtime increase.

\item NOGD has lower accuracy than OGD on seven out of eleven datasets because it employs an approximate feature map of the Laplacian kernel. As a consequence, NOGD can be significantly worse than OGD. Examples are smallNORB, cifar-10, epsilon and mnist. While increasing its budget may improve NOGD's accuracy to approach the level of accuracy of OGD; it will still perform worse than IK$_i$-OGD. Indeed, NODG performed worse than IK$_i$-OGD  on ten out of eleven datasets in Table \ref{tbl:main_results}.

\item IK$_a$-OGD has equal or better accuracy than IK$_i$-OGD. This result is consistent with the assessment comparing the two implementations of Isolation Kernel in density-based clustering \cite{IsolationKernel-AAAI2019}. This is because Voronoi diagram produces partitions of non-axis-parallel regions; whereas iForest yields axis-parallel partitions only. Notice that the accuracy difference between IK$_a$-OGD and OGD is huge on url, news20, rcv1, real-sim and covertype.

\end{itemize}

In terms of runtime:
\begin{itemize}
\item 
While OGD and IK$_i$-OGD are using exactly the same training procedure (with the exception of the prediction function used), IK$_i$-OGD has advantage in two aspects:
\begin{enumerate}
    \item The differences in runtimes are huge---IK$_i$-OGD is three orders of magnitude faster than OGD on seven out of the eleven datasets; and at least one order of magnitude faster on other datasets. This is due to the efficient implementations made possible through Isolation Kernel, described in Section \ref{sec_learning_IK}.
    \item Both OGD and IK$_i$-OGD can potentially incorporate an infinite number of support vectors. But, the prediction function used has denied OGD the opportunity to live up to its full potential because its testing time complexity is proportional to the number of support vectors. In contrast, IK$_i$-OGD has constant test time complexity, independent of the number of support vectors. 
\end{enumerate}

\item Compare with NOGD, IK$_i$-OGD is up to one order of magnitude faster in runtime in high dimensional datasets. On low dimensional datasets (100 or less), IK$_i$-OGD ran only slightly slower. This is remarkable given that IK-OGD has no budget and NOGD has a budget of 100 support vectors only. As a result, NOGD has lower accuracy than IK$_i$-OGD on all datasets, except a9a.

\end{itemize}

In a nutshell, IK-OGD inherits the advantages of OGD (no budget) and NOGD (using $_{primal}f$); yet, it does not have their disadvantages: OGD (using $_{dual}f$); and  NOGD (the need to have a budget which lowers its predictive accuracy).

\subsubsection{Comparison with AdaRaker}
Table \ref{tbl:main_results} shows that multi-kernel learning AdaRaker \cite{OnlineMKL-2018} has lower accuracy than OGD (and even NOGD) using a single kernel. This result is consistent with the previous comparison between SimpleMKL \cite{rakotomamonjy2008simplemkl} and SVM using Isolation Kernel \cite{ting2018IsolationKernel}.  Out of the five datasets on which it could run within reasonable time and without memory errors, AdaRaker ran slower than OGD in three datasets; but faster in two. Compare with IK$_i$-OGD and NOGD, AdaRaker is at least two orders of magnitude slower on the five datasets.
AdaRaker has memory error issues with high dimensional datasets.

\subsubsection{The effects of $t$ on IK-OGD and $b$ on NOGD}
\label{sec_effects}Two datasets, epsilon and smallNORB, are used in this experiment because the accuracy differences between OGD and NOGD on these datasets are the largest; and they are the only two datasets in which IK-OGD performed significantly worse than OGD. We examine the effects of parameters $t$ and $b$ on 
IK-OGD and NOGD.

\begin{figure}[!t]
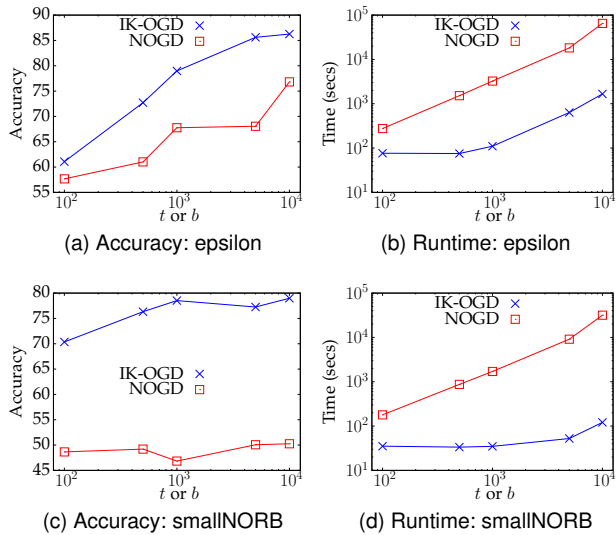

 \centering
\vspace{-12pt}
 \subfloat[Accuracy: epsilon]{\includecombinedgraphics[scale=0.325,vecfile=./epsilon_t_accuracy_v2_pdf]{./epsilon_t_accuracy_v2}}
 \subfloat[Runtime: epsilon]{\includecombinedgraphics[scale=0.325,vecfile=./epsilon_t_timing_v2_pdf]{./epsilon_t_timing_v2}} \\
 \subfloat[Accuracy: smallNORB]{\includecombinedgraphics[scale=0.325,vecfile=./smallNORB_t_accuracy_v2_pdf]{./smallNORB_t_accuracy_v2}}
 \subfloat[Runtime: smallNORB]{\includecombinedgraphics[scale=0.325,vecfile=./smallNORB_t_timing_v2_pdf]{./smallNORB_t_timing_v2}} 
  \caption{Experiments with 
  increasing $t$ for IK$_i$-OGD; and increasing $b$ for NOGD (both Isolation and Laplacian kernels use $\psi=8$). 
  }
  \label{fig_varying_psi_t}
\end{figure}

Figure \ref{fig_varying_psi_t} shows that IK-OGD's accuracy is improved significantly as $t$ increases. Note that, using $t=10000$ on epsilon, the accuracy of IK-OGD reached the same level of accuracy of OGD shown in Table \ref{tbl:main_results}; yet, IK-OGD still ran two orders of magnitude faster than OGD. In contrast, although NOGD's accuracy has improved when $b$ was increased from 100 to 10000, it still performed worse than OGD and IK-OGD by a large margin of 10\%. In addition, NOGD at $b=10000$ ran two orders of magnitude slower than NOGD $b=100$. On smallNORB, IK-OGD also improves its accuracy as $t$ increases up to $t=1000$; but NOGD has showed little improvement over the entire range between $b=100$ and $b=10000$.

NOGD's runtime increases linearly wrt $b$; whereas the runtime of IK-OGD increases sublinearly wrt $t$.

\subsection{CPU and GPU versions of IK$_a$-OGD}
\label{sec_GPUruntime}
The use of Voronoi diagram to partition the data space for Isolation Kernel has slowed down the runtime significantly, compared to that implemented using iForest, mainly due to the need to search for nearest  neighbours. However, because the search for nearest neighbours is amenable to GPU accelerations, we investigate a runtime comparison of the CPU and GPU versions of IK$_a$-OGD.

The result is shown in Table \ref{tbl:GPU_results}.
The GPU version of IK$_a$-OGD is up to four orders of magnitude faster than the CPU version. Despite this GPU speedup, IK$_a$-OGD is still up to one order of magnitude slower than IK$_i$-OGD ran on CPU on some datasets.

In summary, GPU is a good means to speed up IK$_a$-OGD. When accuracy is paramount, IK$_a$-OGD is always a better choice than IK$_i$-OGD (as shown in Table \ref{tbl:main_results}) though the former, even with GPU, runs slower than the latter with CPU.

Note that while it is possible to speed up the original OGD which employs the dual prediction function using GPU, it is not a good solution for two reasons. First, it does not improve OGD's accuracy if the same data independent kernel is used. Second, the GPU-accelerated OGD is expected to still run slower than the CPU version of OGD which employs the primal prediction function using the same kernel. 

The runtime reported in Table \ref{tbl:GPU_results} consists of two components: feature mapping time and OGD runtime. For example, the longest GPU runtime is on epsilon which consists of feature mapping time 457 GPU seconds and OGD runtime of .9 CPU seconds. In other words, the bulk of the runtime is spent on feature mapping; and OGD took only a tiny fraction of a second to complete the job with CPU.

\begin{table}[!t]
 \centering
 \caption{Runtime comparison of the CPU and GPU versions of IK$_a$-OGD\newline (in CPU and GPU seconds, respectively).}
 \label{tbl:GPU_results}
 \begin{tabular}{lrr}
  \toprule
 & {CPU} & {GPU} \\
  \midrule
  url & 1,527 & 65 \\
news20.binary &  1,079 & 10 \\
rcv1.binary & 100,247 & 67\\
real-sim & 31,946 & 10 \\
smallNORB & 406,256 & 178 \\
cifar-10 & 340,047 & 147 \\
epsilon & 1,029,092 & 458 \\
mnist & 56,774 &  45 \\
a9a & 5,589 &  3 \\
covertype & 100,081 & 42 \\
ijcnn1 & 11,999 & 3 \\
  \bottomrule
 \end{tabular}
\end{table}

\subsection{Results with SVM}
\subsubsection{SVM versus IK-SVM}
Without kernel functional approximation,
Isolation Kernel is the only nonlinear kernel, as far as we know, that allows the trick of using $_{primal}f$ 
to be applied to kernel-based methods, including SVM.
We apply Isolation Kernel to SVM to produce IK-SVM.  It is realized using
LIBLINEAR
since IK-SVM is equivalent to applying the IK feature mapped data to a linear SVM. IK-SVM is compared with LIBSVM with Laplacian kernel (denoted as SVM).

\begin{table}[!t]
 \centering
 \caption{SVM versus IK$_a$-SVM. Runtime in CPU seconds.  SVM is LIBSVM with Laplacian kernel; and IK$_a$-SVM is LIBLINEAR with Isolation Kernel. }
 \label{tbl_LIBLINEAR}
 \begin{tabular}{lrr||rr}
  \toprule
  & \multicolumn{2}{c||}{Accuracy} &  \multicolumn{2}{c}{Runtime} \\
 & SVM & IK$_a$-SVM  & SVM & IK$_a$-SVM \\
  \midrule
  url & .67 & .96 & 29,528 & 1.3 \\
news20.binary & .50 & .92 & 684 & 1.5 \\
rcv1.binary & .54 & .96  & 7,472 & .6  \\
real-sim & .75 & .96  & 1,116 & 1.2  \\
smallNORB & --- & .88  & \textcolor{blue}{$>$ 12 hrs} & .9  \\
cifar-10 & .51 & .71 & 3,703 & 1.2  \\
epsilon & --- & .70  & \textcolor{blue}{$>$ 12 hrs} & 90.0  \\
minst & .98 & .99  & 919 & 1.0  \\
a9a & .85 & .84 & 69 & .5  \\
covtype & --- & .93 & \textcolor{blue}{$>$ 12 hrs} & 63.1  \\
ijcnn1 & .99 & .98 & 59 & 2.4  \\
 \bottomrule
 \end{tabular}
\end{table}

Table \ref{tbl_LIBLINEAR} shows the comparison result of SVM and IK$_a$-SVM.
The relative result between SVM and IK$_a$-SVM is reminiscent of that comparing OGD with IK$_a$-OGD in Table~\ref{tbl:main_results}, i.e., IK$_a$-SVM has better accuracy than SVM in all high dimensional datasets; and they have comparable accuracy in datasets less than 2000 dimensions (mnist, a9a and ijcnn1). In terms of runtime, IK$_a$-SVM is up to four orders of magnitude faster. 

Our result in Table \ref{tbl_LIBLINEAR} shows that Isolation Kernel enables SVM to deal with large datasets that would otherwise be impossible practically. 

Note that the runtime reported in Table \ref{tbl_LIBLINEAR} does not include the feature mapping time. With GPU, adding the GPU runtime reported in Table \ref{tbl:GPU_results} (the bulk is the feature mapping time) to that of IK$_a$-SVM does not change the conclusion: IK$_a$-SVM runs order(s) of magnitude faster than SVM and has better accuracy in high dimensional and large scale datasets.

\subsubsection{Compare with additive kernels using SVM}
\label{sec_additive_kernels}

Additive kernels are a class of nonlinear kernels which has approximate feature maps that can be computed efficiently
\cite{AdditiveKernel-PAMI2012, AdditiveKernel-PAMI2013}. These include chi-square and intersection kernels.

They were reported to work well in image datasets when substantial feature engineering such as convolution is performed (e.g., \cite{AdditiveKernel-PAMI2012, AdditiveKernel-PAMI2013}).

Like the Random Fourier Features, the approximate feature maps of additive kernels are generated independent of the given dataset by sampling the continuous spectrum of its Fourier features. The approximate feature map of an additive kernel \cite{AdditiveKernel-PAMI2012} has $2b+1$ features per input dimension, where $b$ is a user-control parameter.
In other words, the feature map has features $2b+1$ times more than the total number of input dimensions. 

One advantage of additive kernels over other kernels is that they have no kernel parameter which needs tuning.

\begin{table}
  \centering
  \setlength{\tabcolsep}{3pt}
  \caption{Kernel versus feature map using SVM for $\chi^2$ additive kernel and Isolation Kernel.  LIBLINEAR uses the feature maps; LIBSVM uses the kernels.\newline \textcolor{green}{-GE-} denotes feature map generation error due to insufficient memory; and it needed more than 64GB to generate the feature map.}
  \label{tbl_additive_kernels}
 \begin{tabular}{lrrcrrcr}
  \toprule
 & \multicolumn{2}{c}{LIBSVM} & & \multicolumn{2}{c}{LIBLINEAR} \\
  \cmidrule{2-3}
  \cmidrule{5-6}
 & IK$_a$ & $\chi^2$ & & $\chi^2$ & IK$_a$ & & $\chi^2$ \#Features \\
  \midrule
url & 0.96 & 0.98 & & \textcolor{green}{-GE-} & 0.96 & & --- \\
news20.binary & 0.89 & 0.97 & & 0.97 & 0.92 & & 1,355,191 \\
rcv1.binary & 0.96 & 0.96 & & 0.95 & 0.96 & & 47,236 \\
real-sim & 0.95 & 0.97 & & 0.97 & 0.96 & & 20,958 \\
smallNORB & 0.88 & 0.81 & & 0.81 & 0.88 & & 18,432 \\
cifar-10 & 0.71 & \textcolor{blue}{$>$ 1 week} & & 0.67 & 0.71 & & 3,072 \\
epsilon & 0.72 & \textcolor{blue}{$>$ 1 week} & & 0.85 & 0.70 & & 2,000 \\
minst & 0.98 & 0.92 & & 0.91 & 0.99 & & 780 \\
a9a & 0.84 & 0.85 & & 0.85 & 0.84 & & 123 \\
covtype & 0.93 & 0.77 & & 0.75 & 0.93 & & 54 \\
ijcnn1 & 0.97 & 0.90 & & 0.91 & 0.98 & & 22 \\
 \bottomrule
 \end{tabular}
\end{table}

Table \ref{tbl_additive_kernels} shows the results of comparing kernel and its feature map using SVM for $\chi^2$ additive kernel and Isolation Kernel. 
Interestingly, the approximate feature map of $\chi^2$ kernel resulted LIBLINEAR to produce accuracies equal or close to those produced by LIBSVM \footnote{The setting $b=0$ is used in the experiment in this section. This appears to be the best setting for the datasets we used here. $b>0$ not only produces poorer accuracy on some datasets, but yields a larger number of features.}. The same result applies to IK$_a$\footnote{LIBSVM appears to perform slightly worse than LIBLINEAR in some datasets with IK$_a$. This is due to the different heuristics used in the optimisation procedures. Otherwise, both LIBSVM and LIBLINEAR shall yield exactly the same accuracy since the exact feature map of Isolation Kernel is used.}.

Note that the method that generates $\chi^2$ feature map has less control on the number of features to be used. On high dimensional datasets, a high number of features must be generated.
This result is consistent with the previous study that one must use very high number of Random Fourier features to achieve good accuracy \cite{Huang-2014}. 

Comparing IK$_a$ with $\chi^2$ with LIBLINEAR, IK$_a$ produced better accuracy than $\chi^2$ in many datasets, especially on dense datasets such as smallNORB, cifar-10, mnist, covtype and ijcnn1. This is despite the fact that IK$_a$ employs $t=100$ features only; and $\chi^2$ employs a lot more features in most datasets. The epsilon dataset is the only dataset in which $\chi^2$ appears to be significantly better than IK$_a$ (85\% vs 70\%); but the accuracy of IK$_a$   could be increased better than 85\% by increasing $t$, as shown in Figure~\ref{fig_varying_psi_t}(a).

In terms of runtime speedup from LIBSVM to LIBLINEAR, both $\chi^2$ and IK$_a$ achieve the same orders of magnitude speedup. For example on the rcv1 dataset, both got 2 and 3 orders of magnitude speedup in training and testing, respectively. But, IK$_a$ enables LIBSVM  to run one order of magnitude faster than $\chi^2$ on the high dimensional rcv1 dataset, i.e., 57 vs 321 seconds and 661 vs 2972 seconds in LIBSVM training and testing, respectively. Using LIBLINEAR, the comparisons are  0.3 vs 0.6 seconds and 0.2 versus 0.7 seconds in training and testing, respectively: IK$_a$ is slightly faster but they are in the same order. The times quoted are   in CPU seconds.

Because of the high number of features generated, the current version of the code\footnote{At \url{scikit-learn.org/stable/modules/kernel\_approximation.html} in Python.} used to generate the $\chi^2$ feature map 
was unable to generate the feature map on the url dataset of over 3 million input dimensions using a machine with 64 GBytes.

We have attempted intersection additive kernel; and it has approximately the same accuracy as the $\chi^2$ kernel.

In summary, while the feature map of $\chi^2$ additive kernel can be approximated well to maintain the accuracy of SVM model achieved by using the kernel, the key weakness is that the number of features cannot be controlled to a manageable number, especially in high dimensional datasets. In addition, it has inferior predictive accuracy in dense datasets in comparison with IK$_a$ in our evaluation.

\section{Relation to existing approaches for efficient kernel methods }
\label{sec_relation}

\subsection{Kernel functional approximation}
Kernel functional approximation is a popular effective approach to produce a user-controllable, finite-dimensional, approximate feature map of a kernel having infinite number of features.

One representative is the Nystr\"{o}m method \cite{Nystrom-NIPS12, Nystrom_NIPS2000,Nystrom_predictive2017}. It first samples $b<n$ points  from the given dataset, and then constructs a matrix of low rank $r$, and derives a vector representation of data of $r$ features. This gives $\mathcal{H}_N = span(\varphi_1, \dots, \varphi_r)$, where $\varphi$ is a normlised eigenfunction of $\sum^b_{i=1} K(\cdot, {\bf x}_i) f({\bf x}_i)$. See \cite{Nystrom-NIPS12} for details. For $r\ll n$, it reduces the search space significantly. 

The key overhead is the eigenvalue decomposition computation of the low rank matrix.
This overhead is not large  only if both $b$ and $r$ are small, relative to the data size $n$ and dimensionality $d$. 
The overhead becomes impracticably large for problems which require large $b$ and $r$.  

Also, though the Nystr\"{o}m method depends on data when deriving an approximate feature map of a chosen nonlinear kernel, but the kernel it is approximating is still data independent (e.g., Gaussian and Laplacian kernels). %

The second representative is random features method \cite{RandomFeatures2007,RandomLaplaceFeatures}. It generates a proxy of features through some transform (such as Fourier or Laplace transform) of the chosen nonlinear shift-invariant kernel function. Only a random subset of these features are used as the feature map. Note that these features are generated independent of the given dataset\footnote{One study has attributed the data independence of the feature generation as the reason of poorer SVM accuracy in comparison with the Nystr\"{o}m method \cite{Nystrom-NIPS12}.}. Let $s$ be the number of random features generated. This gives $\mathcal{H}_R = span(\vartheta_1, \dots, \vartheta_s)$. For $s\ll n$, it reduces the search space significantly. This method has high space complexity which requires to store a $s \times d$ matrix for random Fourier features computations; thus it is not suitable when $d$ is high. A study reported that the number of random features needs to be in order of 200,000 to achieve acceptable accuracy in an application \cite{Huang-2014}. There are faster versions, e.g., FastFood \cite{FastFood-ICML-2013}; but the improved speed often trades off accuracy. The reverse is true for a method which produces a better approximation e.g., \cite{musco2017recursive}.

In contrast with the Nystr\"{o}m method, the random features methods do not need to use a dataset or a sample in the approximation process. In other words, they need no budget. But, a comparison using OGD \cite{LargeScaleOnlineKernelLearning} has shown that the Fourier Random features version always produced lower accuracy than the Nystr\"{o}m version  even when the former used four times more features than the latter in batch mode;  and it could run slower in training and testing.

In a nutshell, the efficiency gain from the kernel functional approximation approach comes with the cost of reduced accuracy as it is an approximation of the chosen nonlinear kernel function.

In contrast, Isolation Kernel has an {\em exact} feature map. As a result, the efficiency gain from the use of Isolation Kernel does not degrade accuracy. It is a {\em direct} method which does not need an intervention step to approximate a feature map from a kernel having infinite or large number of features.

\subsection{Sparse kernel approximation}
To represent non-linearity, the feature map  of a kernel has dimensionality which is usually significantly larger than the dimension of the given dataset. The Nystr\"{o}m method reduces the dimensionality to produce a dense representation. 

In contrast, sparse kernel approximation  aims to produce high-dimensional sparse features\footnote{A sparse representation yields vectors having many zero values, where a feature with zero value means that the feature is irrelevant.}. One proposal \cite{SparseRepresentation2012} approximates each feature vector of ${\bf x}$ using a small subset of representative points, e.g., ${\bf x}$'s neighbours (rather than all representative points). It then uses product quantization (PQ)\footnote{Product Quantization (an improvement over vector quantization) aims to reduce storage  and retrieval time for conducting approximate nearest neighbour search.} %
to encode the sparse features,
and employ bundle methods to learn directly from the PQ codes.

Interestingly, each feature vector of ${\bf x}$ of
Isolation Kernel is both a sparse representation and a coding which employs exactly $t$ representative points, 
from $t$ random subsets of $\psi$ points, i.e., exactly one out of the $\psi$ points in one subset is used for the sparse representation, concatenated $t$ times.  

There are other sparse representations, e.g., (a) Local Deep Kernel Learning \cite{LocalDeepKernelLearning-ICML-2013}
learns a tree-based feature embedding which is high dimensional and sparse through a generalised version of Localized Multiple Kernel Learning of multiple data independent kernels. (b) Concomitant Rank Order (CRO) kernel \cite{CROification-PAMI2019} approximates the Gaussian kernel on the unit sphere. It uses Discrete Cosine Transform to compute the random projection of CRO feature map which can produce feature vectors efficiently.

The key difference between Isolation Kernel and current sparse kernel approximation is that the former is a {\em data dependent} kernel \cite{ting2018IsolationKernel,IsolationKernel-AAAI2019} which has an {\em exact} feature map.
Sparse kernel approximation may be viewed as another intervention step (alternative to kernel functional approximation) to produce a finite-dimensional sparse {\em approximate} feature map from one or more {\em data independent} kernels having infinite number of features. In addition, computationally expensive learning \cite{LocalDeepKernelLearning-ICML-2013} or PQ \cite{SparseRepresentation2012} are not required in Isolation Kernel.

\section{Discussion}

It is important to note that Isolation Kernel is not one kernel function such as Gaussian kernel, but a class of kernels which has different kernel distributions depending on the space partitioning mechanism employed. We use two implementations of Isolation Kernel: (a) iForest \cite{liu2008isolation} which has its kernel distribution similar to that of Laplacian Kernel under uniform density distribution \cite{ting2018IsolationKernel}; (b) when a Voronoi diagram is used to partition the space \cite{IsolationKernel-AAAI2019}, Isolation Kernel has its distribution more akin to an exponential kernel under uniform density distribution. Both realisations of Isolation Kernel adapt to local density of a given dataset, unlike existing data independent kernels. The criterion required of a partitioning mechanism in order to produce an effective Isolation Kernel is described in \cite{ting2018IsolationKernel,IsolationKernel-AAAI2019}. This paper has focused on efficient implementations of Isolation Kernel in online kernel learning, without compromising accuracy.

When using linear kernel, the trick of using $_{primal}f$ instead of $_{dual}f$ to speed up the runtime of both the training stage and the testing stage has been applied previously, e.g., in LIBLINEAR \cite{LIBLINEAR}, even when it is solving the dual optimisation problem. This is possible in LIBLINEAR because linear kernel has an exact and finite-dimensional feature map. But, if you are using an existing nonlinear kernel such as Gaussian or Laplacian kernel, such a trick cannot be applied to SVM because its feature map is not finite.

Note that the work reported in \cite{LargeScaleOnlineKernelLearning}, including OGD and NOGD, and IK-OGD used here do not address the concept change issue in online setting. Nevertheless, all these works address the efficiency issue in online setting which serves as the foundation to tackling the efficacy issue of large scale online kernel learning under concept change.

Geurts et. al. \cite{geurts2006extremely} describe a kernel view of Extra-Trees (a variant of Random Forest \cite{RandomForests}) where its feature map is also sparse and similar to the one we presented here. However, like Random Forest (RF) kernel \cite{Breiman2000}, this kernel was offered as a view point to explain the behaviour of Random Forest; and no evaluation has been conducted to assess its efficacy using a kernel-based method. Ting et. al. \cite{ting2018IsolationKernel} have provided the conceptual differences between RF-like kernels and Isolation Kernel; and their empirical evaluation has revealed that RF-like kernels are inferior to Isolation Kernel when used in SVM.

\section{Concluding remarks}

We began our investigation in questioning the assumption of current approaches to large scale online kernel learning, i.e., the kernel used has a feature map with intractable dimensionality. While this is true for most existing kernels, we reveal that there is one recent kernel called Isolation kernel that has an {\em exact, sparse and finite-dimensional} feature map. 

The new feature map becomes the heart of the proposed approach  to large scale online kernel learning. Using this new approach with Isolation Kernel, we show that large scale online kernel learning can be achieved efficiently without sacrificing accuracy.
It has enabled kernel learning to achieve the outcome which has evaded current approaches thus far, i.e., to live up to its full potential in online setting with large scale high dimensional sparse and dense datasets.

Isolation Kernel's exact, sparse and finite-dimensional feature map is the crucial factor that bring about this outcome. Specifically,  the  proposed feature map enables three key elements: (i) kernel learning with exact finite-dimensional feature map; (ii) sparse representation enables efficient dot product; and (iii) the aNNE implementation of Isolation Kernel is amenable to GPU acceleration.

The proposed approach is generic in two aspects. First, it is not restricted to Isolation Kernel only. Potentially, any data dependent kernel which has an exact, sparse and finite-dimensional feature map can use this approach. Second, even restricting to Isolation Kernel only, as long as a new space partitioning mechanism that can produce a feature map of such properties, it can use this approach as well. 

Future investigation following this approach can focus solely on efficient issues without worrying about the accuracy degradation---the distinguishing feature of this approach over existing approaches.
\bibliography{references}
\bibliographystyle{IEEEtran}

\end{document}